\documentclass{article}
\usepackage{amssymb}
\usepackage{xcolor}
\usepackage[implicit=false]{hyperref}

\long\def\comment#1{}

\newtheorem{assumption}{Assumption}
\newtheorem{theorem}{Theorem}

\newtheorem{corollary}[theorem]{Corollary}
\newtheorem{lemma}[theorem]{Lemma}

\newcommand{\Ass}[1]{Assumption~\ref{#1}}

\newcommand{\Eq}[1]{Equation~\ref{#1}}

\newcommand{\qed}{\mbox{$\Box$}}

\newcommand{\X}{X}
\newcommand{\x}{x}
\newcommand{\Db}{d}
\newcommand{\U}{{\bf X}}
\newcommand{\C}{{\bf x}}
\newcommand{\bY}{{\bf Y}}
\newcommand{\by}{{\bf y}}
\newcommand{\Pai}{{\bf Pa}_i}
\newcommand{\pai}{{\bf pa}_i}
\newcommand{\pail}{{\bf pa}_{il}}
\newcommand{\PaSi}{{\bf Pa}^m_i}
\newcommand{\paSi}{{\bf pa}^m_i}
\newcommand{\PaScii}{{\bf Pa}^{m_{ci}}_i}

\comment{
\newcommand{\U}{{X}}
\newcommand{\C}{{x}}
\newcommand{\bY}{{Y}}
\newcommand{\by}{{y}}
\newcommand{\Pai}{{Pa}_i}
\newcommand{\pai}{{pa}_i}
\newcommand{\pail}{{pa}_{il}}
\newcommand{\PaSi}{{Pa}^m_i}
\newcommand{\paSi}{{pa}^m_i}
\newcommand{\PaScii}{{Pa}^{m_{ci}}_i}

}

\newcommand{\Bm}{m}
\newcommand{\Bmc}{m_c}
\newcommand{\Bmci}{m_{ci}}

\newcommand{\Bmone}{m_1}
\newcommand{\Bmtwo}{m_2}

\newcommand{\Bs}{s}
\newcommand{\Bsc}{s_c}
\newcommand{\Bsci}{s_{ci}}

\newcommand{\Bsone}{s_1}
\newcommand{\Bstwo}{s_2}

\newcommand{\hBs}{m^h}
\newcommand{\hBsc}{m^h_c}
\newcommand{\hBsci}{m^h_{ci}}

\newcommand{\hBsone}{m^h_1}
\newcommand{\hBstwo}{m^h_2}

\newcommand{\Fs}{{\cal F}_s}
\newcommand{\Fsone}{{\cal F}_{s1}}
\newcommand{\Fstwo}{{\cal F}_{s2}}

\newcommand{\Jarai}{J\'{a}rai }
\newcommand{\Aczel}{Acz\'{e}l }

\newcommand{\dXdY}[2]{\partial {#1} / \partial {#2}}

\newcommand{\Th}{\mbox{$\theta$}}

\newcommand{\Thi}{\Th_i}
\newcommand{\TBs}{\Th_{m}}
\newcommand{\TBsc}{\Th_{mc}}
\newcommand{\TBsci}{\Th_{mci}}
\newcommand{\TBsone}{\Th_{m1}}
\newcommand{\TBstwo}{\Th_{m2}}

\newcommand{\CTBs}{\Theta_{m}}
\newcommand{\CTBsone}{\Theta_{m1}}

\newcommand{\vecm}{{m}}
\newcommand{\vecv}{{v}} 
\newcommand{\vecb}{{b}} 
\newcommand{\matW}{W} 
\newcommand{\matS}{S} 
\newcommand{\matM}{M} 
 
\newcommand{\Xvecbar}{\overline{{\bf x}}}
\newcommand{\Xvecsubl}{{\bf x}_{\textcolor{blue}{l}}}
\newcommand{\vecmu}{\mu}
\newcommand{\matB}{B} 
\newcommand{\tr}{\mbox{\it tr}} 
\newcommand{\amu}{\alpha_{\mu}} 
\newcommand{\aw}{\alpha} 
\newcommand{\awY}{\alpha'}
 
\newcommand{\A}{W_{11}} 
\newcommand{\B}{W_{12}} 
\newcommand{\D}{W_{22}} 

\title{Parameter Priors for Directed Acyclic Graphical 
Models and the Characterization of Several Probability 
Distributions}

\author{
Dan Geiger\\ 
geiger02@gmail.com\\
\\
David Heckerman\\
heckerma@hotmail.com}

\date{July 1999, Revised June 2021}

\begin{document}

\maketitle

\begin{abstract}
\noindent
We show that the only parameter prior
for complete Gaussian DAG models 
that satisfies global parameter independence,
complete model equivalence, and some weak regularity
assumptions, is the normal-Wishart distribution.  
Our analysis is based on the
following new characterization of the Wishart distribution:
let $W$ be an $n \times n$, $n \ge 3$, positive-definite symmetric 
matrix of random variables
and $f(W)$ be a pdf of $W$.
Then, f$(W)$ is a Wishart distribution if and only if
$W_{11} - W_{12} W_{22}^{-1} W'_{12}$ is independent of 
$\{W_{12},W_{22}\}$
for every block partitioning $W_{11},W_{12}, W'_{12}, W_{22}$ of
$W$. 
Similar characterizations of the normal and normal-Wishart
distributions are provided as well.
We also show how to construct a prior for every DAG model
over $\U$ from the prior of a single regression model.

\comment{
\noindent We develop simple methods for constructing
parameter priors for model choice among
Directed Acyclic Graphical (DAG) models.  
In particular, we introduce several
assumptions that permit the construction of parameter
priors for a large number of DAG models from a small
set of assessments.  
We then present a method for directly computing the
marginal likelihood of every DAG model given
a random sample with no missing observations.
We apply this methodology to Gaussian DAG models 
which consist of a recursive set of linear regression models.  
We show that the only parameter prior
for complete Gaussian DAG models 
that satisfies our assumptions
is the normal-Wishart distribution.  
Our analysis is based on the
following new characterization of the Wishart distribution:
let $W$ be an $n \times n$, $n \ge 3$, positive-definite symmetric 
matrix of random variables
and $f(W)$ be a pdf of $W$.
Then, f$(W)$ is a Wishart distribution if and only if
$W_{11} - W_{12} W_{22}^{-1} W'_{12}$ is independent of 
$\{W_{12},W_{22}\}$
for every block partitioning $W_{11},W_{12}, W'_{12}, W_{22}$ of
$W$. 
Similar characterizations of the normal and normal-Wishart
distributions are provided as well.

\vspace{1ex}
} 

\bigskip

\noindent
Corrections to the original text in \textcolor{red}{red} are taken
from J. Kuipers, G. Moffa, and D. Heckerman, Addendum on the scoring
of Gaussian directed acyclic graphical models. {\em Annals of Statistics} 
42, 1689-1691, Aug 2014. 
Other updates to the original are in \textcolor{blue}{blue}.

\end{abstract}

\section{Introduction} \label{sec:intro}

Directed Acyclic Graphical (DAG) models
have increasing number of applications in Statistics
(Spiegelhalter, Dawid, Lauritzen, and Cowell, 1993)
\nocite{SDLC93} 
as well as in Decision Analysis 
and Artificial Intelligence 
(Heckerman, Mamdani, Wellman, 1995b; Howard and Matheson, 1981;
Pearl, 1988). \nocite{Howard81,Pearl88,HMW95cacm}
A DAG model $\Bm = (\Bs, \Fs)$ 
for a set of variables $\U=\{X_1,\ldots,X_n\}$
each associated with a set of possible values $\mbox{D}_i$, respectively,
is a set of joint probability distributions for 
$\mbox{D}_{1} \times \cdots \times \mbox{D}_{n}$ 
specified via two components: a structure $\Bs$ and a set of local distribution
families $\Fs$. The structure $\Bs$ for $\U$ is a directed graph
with no directed cycles (i.e., a Directed Acyclic Graph)
having for every variable $X_i$ in $\U$
a node labeled $X_i$ with parents labeled by $\PaSi$. 
The structure $\Bs$ represents the set of 
conditional independence assertions,
and only these conditional independence assertions, 
which are implied by
a factorization of a joint distribution for $\U$ given by
$p(\C) = \prod_{i=1}^n p(\x_i|\paSi)$,
where $\C$ is a value for $\U$ (an $n$-tuple) and
$x_i$ is a value for $X_i$.  When $\x_i$ has no incoming
arcs in $m$ (no parents), $p(\x_i|\paSi)$ stands for $p(\x_i)$.
The local distributions are the $n$ conditional and marginal
probability distributions that constitute the factorization
of $p(\C)$. Each such distribution belongs to the
specified family of allowable probability distributions $\Fs$.
A DAG model is often called a {\em Bayesian network}, although
the later name sometimes refers to a specific joint probability 
distribution that factorizes according to a DAG, and 
not, as we mean herein, a set of joint distributions each factorizing 
acccording to the same DAG.
A DAG model is {\em complete} if it has no missing
arcs. Note that any two complete DAG models for $\U$
encode the same assertions of conditional independence, 
namely none.

In this paper, we assume that
each local distribution is selected from 
a family $\Fs$ which depends on a
finite set of parameters $\TBs \in \CTBs$
(a parametric family). The parameters for a local
distribution is a set of real numbers that 
completely determine the functional form of
$p(\x_i|\paSi)$ when $x_i$ has parents
and of $p(\x_i)$ when $x_i$ has no parents.
We denote by $\hBs$ the model hypothesis that the 
true joint probability distribution of $\U$ 
is perfectly represented by a structure $\Bs$ of a DAG model $\Bm$
with local distributions from $\Fs$, namely, 
that the joint probability distribution satisfies 
only the conditional independence assertions
implied by this factorization and none other.
Consequently, the true joint distribution for a DAG 
model $\Bm$ is given by,
\begin{equation} \label{eq:bn-def-p}
p(\C|\TBs,\hBs) = \prod_{i=1}^n p(\x_i|\paSi,\Thi,\hBs)
\end{equation}
where $\by=\{\x_i\}_{\X_i \in \bY}$
denotes a value of $\bY \subseteq \U$
and $\theta_1,\ldots \theta_n$ are subsets of $\TBs$.
Whereas in a general formulation of DAG models,
the subsets $\{\theta_i\}_{i=1}^n$ could possibly ovelap 
allowing several local distribution to have common parameters,
in this paper, we shall shortly exclude this possibility
(Assumption~\ref{ass:pi}). Note that $\theta_m$ denotes the
union of $\theta_1, \ldots,\theta_n$ for a DAG model $m$.

We consider the Bayesian approach when
the parameters $\TBs$ and the
model hypothesis $\hBs$ are uncertain
but the parametric families are known.
Given data
$\Db=\{\C_1,\ldots,\C_m\}$, a random sample from $p(\C|\TBs,\hBs)$
where $\TBs$ and $\hBs$ are the true parameters and model
hypothesis, respectively, we can compute the posterior probability of
a model hypothesis $\hBs$ using

\begin{equation} \label{eq:post-bs}
p(\hBs|\Db) = c \ p(\hBs) \ p(\Db|\hBs) =
c \ p(\hBs) \int p(\Db|\TBs,\hBs) \ p (\TBs|\hBs) \ d \TBs
\end{equation}

where $c$ is a normalization constant.
We can then select a DAG model
that has a high posterior probability or average
several good models for prediction.  

The problem of selecting an appropriate DAG model,
or sets of DAG models, given data, posses
a serious computational challenge, 
because the number of DAG models grows faster than exponential
in $n$.  Methods for searching through the
space of model structures are discussed (e.g.) 
by Cooper and
Herskovits (1992), Heckerman, Geiger, and Chickering
(1995a), and Friedman and Goldszmidt (1997).
\nocite{CH92,HGC95ml,FG97}

From a statistical viewpoint, an important question
which needs to be addressed is how to specify the quantities
$p(\hBs)$, $p(\Db|\TBs,\hBs)$, $p (\TBs|\hBs)$, 
needed for evaluating $p(\hBs|\Db)$
for every DAG model $\Bm$ that could 
conceivably be considered by a search algorithm.
Buntine (1991) and Heckerman et al. (1995a) discuss methods for
specifying the priors $p(\hBs)$ 
via a small number of direct assessments.
Geiger and Heckerman (1994) and Heckerman and Geiger (1995)
\nocite{GH94uai,HG95uai}
develop practical methods for assigning 
parameter priors $p (\TBs|\hBs)$ to every candidate DAG
model $\Bm$ via a small number of direct assessments. 
Another relevant paper is by
Dawid and Lauritzen (1993) \nocite{DL93}
who discuss the notion of hyper and meta markov laws.

\comment{
Our method is based
on a set of assumptions the most notable of which 
is the assumption that complete DAG models represent
the same set of distributions, which implies that
data cannot distinguish between two complete DAG models.
Multivariate Gaussian, multinomial, and multivariate 
$t$ distributions satisfy this assumption.
Another assumption is
{\em likelihood and prior modularity}, 
which says that the local distribution for $x_i$
and its parameter priors
depend only on the parents of $x_i$ but
not on the entire description of the structure.
These assumptions, together with 
{\em global parameter independence} 
(Dawid and Lauritzen, 1993)\nocite{DL93}, 
are the heart of the proposed methodology.
}

The contributions of this paper are twofold:
A methodology for specifying parameter priors for
Gausian DAG models using a prior for a single
regression model (Section~2).
An analysis of complete Gaussian DAG models 
which shows that the only parameter prior
that satisfies our assumptions
is the normal-Wishart distribution (Section~3).

The analysis is based on the
following new characterization of the 
Wishart, normal, and normal-Wishart distributions.

\comment{
let $W$ be an $n \times n$, $n \ge 3$, positive-definite symmetric 
matrix of random variables
and $f(W)$ be a pdf of $W$.
Then, f$(W)$ is a Wishart distribution if and only if
$W_{11} - W_{12} W_{22}^{-1} W'_{12}$ is independent of 
$\{W_{12},W_{22}\}$
for every block partitioning $W_{11},W_{12}, W'_{12}, W_{22}$ of
$W$. 
Similar characterizations of the normal and normal-Wishart
distributions are provided as well (Section~3). 
}

\vspace{1ex}

\noindent
{\bf Theorem}
{\em 
Let $W$ be an $n \times n$, $n \ge 3$, positive-definite symmetric 
matrix of real random variables such that no entry in $W$ is zero,
$\mu$ be a an $n$-dimentional vector of random variables,
$f_W(W)$ be a pdf of $W$,
$f_{\mu}(\mu)$ be a pdf of $\mu$,
and $f_{\mu,W}(\mu,W)$ be a pdf of $\{\mu,W\}$.
Then, 
$f_W(W)$ is a Wishart distribution,
$f_{\mu}(\mu)$ is a normal distribution,
and $f_{\mu,W}(\mu,W)$ is a normal-Wishart distribution
if and only if global parameter independence holds
for unknown $W$, unknown $\mu$, or unknown $\{\mu,W\}$,
respectively.
}

The assumption of global parameter independence
is expressed differently for each of the three cases
treated by this theorem and the 
proof follows from
Theorems~\ref{thm:Wishart},~\ref{thm:normal} 
and~\ref{thm:normalWishart}, respectively,
proven in Section~3.
It should be noted that a single principle, global parameter
independence,
is used to characterizes three different distributions.
In Section~4, we
compare these characterizations to a recent 
characterization of the Dirichlet distribution
(Geiger and Heckerman, 1997; \Jarai, 1998) \nocite{GH97stat,Ja98}
and conjecture that the later characterization 
uses a redundant assumption 
(local parameter independence)---that is, global parameter 
independence may also characterize
the Dirichlet distribution.  The Dirichlet, normal, Wishart, 
and normal-Wishart distributions are the
conjugate distributions for the standard 
multivariate exponential families.

\section{Priors for DAG models}

In this section we provide a novel presentation 
of our previous results in 
(Geiger and Heckerman, 1994; Heckerman and Geiger, 1995).
\nocite{GH94uai,HG95uai}
We have sharpenned the 
assumptions involved in learning DAG models
with no hidden variables from
complete data.  As a result, we show that
a prior for one regression model dictates, under
our assumptions, the prior for all Gaussian DAG models
over the same variables.  Our new presentation, which uses
matrix notation for expressing independence of parameters
of Gaussian DAG models, enables us to prove the
characterization theorems in the next section.

This section is organized as follows:
A methodology for specifying parameter priors for
many structures using a few direct assessments
(Section~2.1).
A formula that computes the marginal likelihood
for every dag model (Section~2.2). 
A specialization of this formula
to an efficient computation for
Gaussian DAG models (Section~2.3).

\subsection{The Construction of Parameter Priors} 
\label{sec:like-pri}

We start by presenting a set of assumptions that simplify the assessment
of parameter priors and a method of assessing these priors.
The assumptions are as follows:

\begin{assumption}[Complete model equivalence] \label{ass:cme}
Let $\Bmone = (\Bsone, \Fsone)$ be a complete DAG model
for a set of variables $\U$.
The family $\Fstwo$ of every complete DAG model
$\Bmtwo = (\Bstwo, \Fstwo)$ for $\U$ is such that
$\Bmone$ and $\Bmtwo$ 
represent the same set of joint probability distributions.
\end{assumption}

We explain this assumption by providing an example where
it fails.
Suppose the set of variables $\U=\{\X_1,\X_2,\X_3\}$ consists of 
three variables each with possible 
values $\{x_i, \overline{x}_i\}$, respectively, and
$\Bsone$ is the complete structure with arcs
$\X_1 \rightarrow \X_2$, $\X_1 \rightarrow \X_3$, and $\X_2
\rightarrow \X_3$. 
Suppose further, that
the local distributions $\Fsone$ of model $\Bmone$ 
are restricted to the sigmoid function
\begin{displaymath}
p(\x_i|\paSi,\Thi,\hBs)= \frac{1}{
  1 + {\rm exp}\left\{ a_i + \sum_{\x_j \in \paSi} b_{ji} \x_j \right\}}
\end{displaymath}
where $\theta_1= \{a_1\}$,
$\theta_2= \{a_2, b_{12}\}$,
and $\theta_3 = \{a_3, b_{13}, b_{23}\}$.

Consider now a second complete model $\Bmtwo$ for $\U=\{\X_1,\X_2,\X_3\}$
whose structure consists of the arcs $\X_1 \rightarrow \X_2$, 
$\X_1 \rightarrow \X_3$, and $\X_3 \rightarrow \X_2$.  
Assumption~\ref{ass:cme} asserts that the families of 
local distributions for $\Bmone$ and $\Bmtwo$ are such that
the set of joint distributions for $\U$ represented
by these two complete models is the same.  
In this example, however, if we specify the local families
for $\Bmtwo$ by also restricting
them to be sigmoid functions,
the two models will represent different sets of
joint distributions over
$\{\X_1,\X_2,\X_3\}$. Hence,
Assumption~\ref{ass:cme} will be violated.
Using Bayes rule one can always determine a 
set of local distribution families that
will satisfy Assumption~\ref{ass:cme}, however,
their functional form will usually involve an integral
(and will often violate Assumption~\ref{ass:pi} below).
A notable exception is discussed in Section~\ref{sec:linear}.

Our definition of $\hBs$, that the true
joint pdf of a set of variables $\U$ is perfectly represented by $m$,
and Assumption~\ref{ass:cme}, which says that two complete
models represent the same set of joint pdfs for $\U$, 
imply that for two complete models $\hBsone = \hBstwo$.
This is a strong assumption.
It implies that 
$p(\TBstwo|\hBstwo) = p(\TBstwo|\hBsone)$
because two complete models represent the same set of distributions.  
It also implies $p(\Db|\hBsone) = p(\Db|\hBstwo)$ which says
that the marginal likelihood for two complete DAG models is
the same for every data set, or equivalently, that complete
DAG models cannot be distinguished by data.
Obviousely, in the example with the sigmoid functions,
the two models can be distinguished by data 
because they do not represent
the same set of joint distributions.
\footnote{
A technical point worth mentioning here is our use of the term
variable and its relationship to the standard definition
of a random variable.  A {\em continuous random variable} $X$, 
according to most probability text books,
is a function $X: \Omega \rightarrow R$ such that
$\{w | X(w) \leq x\} \in \cal A$ where 
$\cal A$ is a $\sigma$-field of subsets of $\Omega$ and
$\Omega$ is a sample space
of a probability space $(\Omega, {\cal A}, P)$ and where
$P$ is a fixed probability measure.
A {\em discrete random variable}
is a function $X: \Omega \rightarrow D$ where $D$ is a discrete set
such that $\{w | X(w) = x_i \} \in \cal A$ for every $x_i \in D$
where $\cal A$ is a $\sigma$-field and
$\Omega$ is a sample space
of a probability space $(\Omega, {\cal A}, P)$.
We use the term {\em variable},
as common to much of the literature on DAG models, 
to mean a function $X_i: \mbox{\cal A} \rightarrow D_{i}$,
where {\cal A} is a $\sigma$-field of subsets of $\Omega$, 
parallel to the usual definition of a random variable,
but without fixing a specific probability measure $P$.
A model $m$ for a set of variables $\U$, 
(and a DAG model in particular), is simply a set of 
probability measures on the Cartesian product $\times_i D_{i}$.  
Once a particular probability measure from $m$ is picked, a variable
in our sense becomes a random variable in the usual sense.
}

\comment{OLD alternative
A standard definition of a {\em model} is a set of probability
measures.
One cannot speak about a model for a set of random variables $\U$,
as we need here, because, according to standard definitions, 
a random variable already has a specific probability 
measure $P$.  Our use of the word variable, common to much
of the literature on DAG models, circumvents this 
difficulty by defining a variable $X_i$ to be, 
a function $X_i: \mbox{\cal A} \rightarrow D(X_i)$,
where {\cal A} is a $\sigma$-field of subsets of $\Omega$, 
parallel to the usual definition of a random variable,
but without fixing a specific probability measure $P$.
Now a model over a set of variables $\U$, 
(and a DAG model in particular), is simply a set $m$ of 
probability measures on the Cartesian product $\times_i D(X_i)$.  
Once a particular probability measure from $m$ is picked, a variable,
in our sense, becomes a random variable, in the usual sense.
}

\comment{
We will consider all structures for $\X$ to be feasible---that
is, $p(\hBs)>0$ for all $\Bs$ for $\X$.
}

\begin{assumption}[Regularity] \label{ass:regular}
For every two complete DAG models $\Bmone$ and $\Bmtwo$ for $\U$
there exists a one-to-one mapping $f_{12}$
between the parameters $\TBsone$ of $\Bmone$
and the parameters $\TBstwo$ of $\Bmtwo$
such that the likelihoods satisfy
$p(\C|\TBsone,\hBsone)  = p(\C|\TBstwo,\hBstwo) $
where $\TBstwo=f_{1,2}(\TBsone$).
The Jacobian
$|\dXdY{\TBsone}{\TBstwo}|$ exists and is 
non-zero for all values of $\CTBsone$.
\end{assumption}

Assumption~\ref{ass:regular} implies
$p(\TBstwo|\hBsone) =  \left| \frac{\partial \TBsone}{\partial \TBstwo}
\right| \ p(\TBsone|\hBsone)$ where
$\TBstwo=f_{1,2}(\TBsone$).
Furthermore, 
due to Assumption~\ref{ass:cme},
$p(\TBstwo|\hBstwo) = p(\TBstwo|\hBsone)$,
and thus
\begin{equation}
\label{prior_eq}
p(\TBstwo|\hBstwo) =  \left| \frac{\partial \TBsone}{\partial \TBstwo}
\right| \ p(\TBsone|\hBsone).
\end{equation}

\begin{assumption}[Likelihood Modularity] \label{ass:lm}
For every two DAG models $\Bmone$ and $\Bmtwo$ for
$\U$ such that $\X_i$ has the same parents in $\Bmone$ and $\Bmtwo$,
the local distributions for $x_i$ in both models
are the same, namely,
$p(\x_i|\paSi,\Thi,\hBsone) = p(\x_i|\paSi,\Thi,\hBstwo)$
for all $\X_i \in \U$.
\end{assumption}

\comment{
Given Assumptions
\ref{ass:cme}, \ref{ass:regular}, and \ref{ass:lm}, 
it follows that
the likelihood $p(\C|\TBs,\hBs)$
for every DAG model $\Bm$ 
is determined by a specified likelihood 
$p(\C|\TBsc,\hBsc)$
for a single complete DAG model $\Bmc$.
}

\begin{assumption}[Prior Modularity] \label{ass:pm}
For every two DAG models $\Bmone$ and $\Bmtwo$ for
$\U$ such that $\X_i$ has the same parents in $\Bmone$ and $\Bmtwo$,
$p(\Thi|\hBsone) = p(\Thi|\hBstwo)$.
\end{assumption}

\begin{assumption}[Global Parameter Independence] \label{ass:pi}
For every DAG model $\Bm$ for $\U$, 
$p(\TBs|\hBs) = \prod_{i=1}^n p(\Thi|\hBs)$.
\end{assumption}

The likelihood and prior modularity assumptions have 
been used implicitly in the work of (e.g.)
Cooper and Herskovits (1992), Spiegelhalter et al. (1993), 
and Buntine
(1994).\nocite{CH92,SDLC93,Buntine94} 
Heckerman et al.\ (1995a)\nocite{HGC95ml} made \Ass{ass:pm} 
explicit in the context of discrete variables
under the name parameter modularity.
Spiegelhalter and Lauritzen (1990)\nocite{SL90} introduced 
Assumption~\ref{ass:pi} in the context of DAG models
under the name global independence.
Assumption~\ref{ass:pi} excludes the possibility that
two local distributions would share a common parameter.

The assumptions we have made lead to the following 
significant implication:
When we specify a parameter prior $p(\TBsc|\hBsc)$ for one complete 
DAG model $\Bmc$, we also implicitly specify a prior
$p(\TBs|\hBs)$ for any DAG model $\Bm$ among the super 
exponentially many possible DAG models.  Consequently, we have a
framework in which a manageable number of direct assessments
leads to all the priors needed to search the model space.
In the rest of this section, we explicate how all parameter
priors are determined by the one elicited prior.
In Section~2.3, we show how to elicit the one needed prior
$p(\TBsc|\hBsc)$ under specific distributional assumptions.

Due to the complete model equivalence and regularity assumptions, 
we can compute $p(\TBsc|\hBsc)$ for one complete model
for $\U$ from the
prior of another complete model for $\U$.  In so doing, 
we are merely performing coordinate transformations between
parameters for different variable orderings in the factorization of
the joint likelihood (Eq.~\ref{prior_eq}).
Thus by specifying parameter prior for
one complete model, we have implicitly specified a prior
for every complete model.

It remains to examine how the prior
$p(\TBs|\hBs)$ is computed
for an incomplete DAG model $\Bm$ for $\U$.
Due to global parameter independence we have
$p(\TBs|\hBs) = \prod_{i=1}^n p(\Thi|\hBs)$
and therefore it suffices to examine each of the $n$ terms
separately.  To compute
$p(\Thi|\hBs)$, we identify a complete DAG model $\Bmci$
such that $\PaSi=\PaScii$.
The prior $p(\TBsci|\hBsci)$ is obtained from
$p(\TBsc|\hBsc)$, as we have shown for every
pair of complete DAG models.
Now, global parameter independence 
states that $p(\TBsc|\hBsci)$ can be written 
as a product  $\prod_{i=1}^n p(\Thi|\hBsci)$,
and therefore, $p(\Thi|\hBsci)$ is available.
Finally, due to prior modularity 
$p(\Thi|\hBs)$ is equal to $p(\Thi|\hBsci)$.

The following theorem summarizes this discussion.

\begin{theorem} \label{thm:prior}
Given Assumptions
\ref{ass:cme} through
\ref{ass:pi}, 
the parameter prior 
$p(\TBs|\hBs)$ for every DAG model $\Bm$ 
is determined by a specified parameter prior $p(\TBsc|\hBsc)$
for an arbitrary complete DAG model $\Bmc$.
\end{theorem}

Theorem~\ref{thm:prior} shows that once we specify the 
parameter prior for one complete DAG model all other priors
can be generated automatically and need not be specified manually.
Consequently, together with Eq.~\ref{eq:post-bs} and due
to the fact that also likelihoods can be generated
automatically in a similar fashion, we have 
a manageable methodology to automate the computation of
$p(\Db|\hBs)$ for any DAG model of $\U$ which is being
considered by a search algorithm as a candidate model.
Next we show how this computation can be done implicitly 
without actually computing the priors and likelihoods.

\subsection{Computation of the Marginal Likelihood for Complete Data} 
\label{sec:be}

For a given $\U$, consider a DAG model $\Bm$ and a complete random
sample $\Db$.  Assuming global parameter independence, the parameters
remain independent given complete data.  That is,
\begin{equation} \label{eq:post-pi} 
p(\TBs|\Db,\hBs) = 
  \prod_{i=1}^n p(\Thi|\Db,\hBs) 
\end{equation}
In addition, assuming global parameter independence, likelihood
modularity, and prior modularity, the parameters remain modular
given complete data.  In particular, if $\X_i$ has the same parents in
$\Bsone$ and $\Bstwo$, then 
\begin{equation} \label{eq:post-pm} 
p(\Thi|\Db,\hBsone) = p(\Thi|\Db,\hBstwo)
\end{equation}
Also, for any $\bY \subseteq \U$, define $\Db^{\bY}$ to be the random
sample $\Db$ restricted to observations of $\bY$.  For example, if
$\U=\{\X_1,\X_2,\X_3\}$, $\bY=\{\X_1,\X_2\}$, and
$\Db=\{\C_1=\{\x_{11},\x_{12},\x_{13}\},
\C_2=\{\x_{21},\x_{22},\x_{23}\}\}$, then we have $\Db^{\bY} = \{
\{\x_{11},\x_{12}\}, \{\x_{21},\x_{22}\}\}$.  
Let $\bY$ be a subset of $\U$, and $\Bsc$ be a complete structure for
any ordering where the variables in $\bY$ come first.  Then, assuming
global parameter independence and likelihood modularity, it is not
difficult to show that
\begin{equation} \label{eq:ignore}
p(\bY|\Db,\hBsc) = p(\bY|\Db^{\bY},\hBsc)
\end{equation}
Given these observations, we can compute the marginal likelihood as
follows.

\begin{theorem} \label{thm:be}
Given any complete DAG model $\Bmc$ for $\U$, any DAG model $\Bm$ for
$\U$, and any complete random sample $\Db$, Assumptions~\ref{ass:cme}
through \ref{ass:pi} imply
\begin{equation} \label{eq:be}
p(\Db|\hBs) = \prod_{i=1}^n 
  \frac{p(\Db^{\Pai \cup \{\X_i\}}|\hBsc)}{
        p(\Db^{\Pai}|\hBsc)}
\end{equation}
\end{theorem}

\noindent {\bf Proof:}
From the rules of probability, we have

\begin{equation} \label{eq:be1a}
p(\Db|\hBs) = \prod_{l=1}^m \int p(\C_l|\TBs,\hBs) \ p(\TBs|\Db_l,\hBs) 
  \ d\TBs
\end{equation}

\noindent
where $\Db_l = \{\C_1,\ldots,\C_{l-1}\}$.  Using
Equations~\ref{eq:bn-def-p} and \ref{eq:post-pi} to rewrite the first
and second terms in the integral, respectively, we obtain

\begin{displaymath} \label{eq:be1b}
p(\Db|\hBs) = \prod_{l=1}^m \int \prod_{i=1}^n p(\x_{il}|\pail,\Thi,\hBs) \ 
  p(\Thi|\Db_l,\hBs) \ d\TBs
\end{displaymath}
where $\x_{il}$ is the value of $X_i$ in the $l$-th data point.

\noindent
Using likelihood modularity and \Eq{eq:post-pm}, we get

\begin{equation} \label{eq:be2}
p(\Db|\hBs) = \prod_{l=1}^m \int \prod_{i=1}^n p(\x_{il}|\pail,\Thi,\hBsci) \ 
  p(\Thi|\Db_l,\hBsci) \ d\TBs
\end{equation}

\noindent
where $\Bsci$ is a complete structure with variable ordering
$\Pai$, $\X_i$ followed by the remaining variables.  Decomposing
the integral over $\TBs$ into integrals over the individual parameter
sets $\Thi$, and performing the integrations, we have
\begin{displaymath} \label{eq:be3}
p(\Db|\hBs) = \prod_{l=1}^m \prod_{i=1}^n 
  p(\x_{il}|\pail,\Db_l,\hBsci)
\end{displaymath}
Using \Eq{eq:ignore}, we obtain
\begin{eqnarray} \label{eq:be4}
p(\Db|\hBs) & = & \prod_{l=1}^m \prod_{i=1}^n 
  \frac{ p(\x_{il},\pail|\Db_l,\hBsci) }{ p(\pail|\Db_l,\hBsci) }
     \nonumber \\*[9pt]
& = & \prod_{l=1}^m \prod_{i=1}^n 
  \frac{
     p(\x_{il},\pail|\Db^{\Pai \cup \{\X_i\}}_l,\hBsci) }{
     p(\pail|\Db^{\Pai}_l,\hBsci) }
     \nonumber \\*[9pt]
& = & \prod_{i=1}^n 
  \frac{
     p(\Db^{\Pai \cup \{\X_i\}}|\hBsci) }{
     p(\Db^{\Pai}|\hBsci) }
\end{eqnarray}
By the likelihood modularity, complete model equivalence,
and regularity assumptions, we have that
$p(\Db|\hBsci) = p(\Db|\hBsc), i=1,\ldots,n$.
Consequently, for any subset $\bY$ of $\U$, we obtain
$p(\Db^{\bY}|\hBsci) = p(\Db^{\bY}|\hBsc)$ by summing over
the variables in $\Db^{\U \setminus \bY}$.  
Consequently, using \Eq{eq:be4}, we get \Eq{eq:be}. \qed

An important feature of the formula for marginal likelihood
(\Eq{eq:be}), which we now demonstrate,
is that two DAG models that represent 
the same assertions of conditional independence have the same 
marginal likelihood.
We say that two structures for $\U$ are {\em independence
equivalent} if they represent the same assertions of conditional
independence.  Independence equivalence is an equivalence relation,
and induces a set of equivalence classes over the possible structures
for $\U$.  

Verma and Pearl (1990)\nocite{Verma90} provide a simple
characterization of independence-equivalent structures
using the concept of a v-structure.
Given a
structure $\Bs$, a {\em v-structure} in $\Bs$ is an ordered node
triple $(\X_i,\X_j,\X_k)$ where $\Bs$ contains the arcs $\X_i
\rightarrow \X_j$ and $\X_j \leftarrow \X_k$, and there is no arc
between $\X_i$ and $\X_k$ in either direction.
Verma and Pearl show that
two structures for $\U$ are independence equivalent if and
only if they have identical edges and identical v-structures.
This characterization makes it easy to identify independence
equivalent structures.  

An alternative characterization by Chickering \nocite{Ci95}
(1995) is useful for proving our claim 
that independence equivalent structures
have the same marginal likelihood.  
An {\em arc reversal} is a transformation from one
structure to another, in which a single arc between two nodes is
reversed.  An arc between two nodes is said to be {\em covered} if
those two nodes would have the same parents if the arc were
removed.
\begin{theorem}[Chickering, 1995] \label{thm:C95}
Two structures for $\U$ are independence equivalent if and only if
there exists a set of covered arc reversals that transform one
structure into the other.
\end{theorem}
A proof of this theorem can also be found in (Heckerman et~al., 1995a).
We are ready to prove our claim.

\begin{theorem} \label{thm:samelikelihood}
Given Assumptions~\ref{ass:cme} through \ref{ass:pi},
every two independence equivalent 
DAG models have the same marginal likelihood.
\end{theorem}

\noindent {\bf Proof:} 
Theorem~\ref{thm:C95} implies that we can restrict 
the proof to two DAG models that differ 
by a single covered arc. Say the arc is between $X_i$
and $X_j$ and that the joint parents of $X_i$ and $X_j$
are denoted by $\pi$.
For these two models, \Eq{eq:be} differs only
in terms $i$ and $j$. For both models
the product of these terms is 
$ p(\Db^{\pi \cup \{\X_i,X_j\}}|\hBsc)/ 
p(\Db^{\pi}|\hBsc)$.
\qed

The conclusions of Theorems~\ref{thm:be}
and~\ref{thm:samelikelihood} are not justified
when our assumptions are violated. 
In the example of the sigmoid functions, discussed in
the previous subsection,
the structures $\Bsone$ and $\Bstwo$ differ by the
reversal of a covered arc between $\X_2$ and $\X_3$, but, given 
that all local distribution families are
sigmoid, there are certain joint likelihoods that can
be represented by one structure, but not the other,
and so their marginal likelihood is different.

\subsection{Gaussian Directed Acyclic Graphical Models}\label{sec:linear} 

We now apply the methodology
of previous sections to Gaussian DAG models.
A Gaussian DAG model is a DAG model as defined
by Eq~\ref{eq:bn-def-p}, where 
each variable $\X_i \in \U$ is continuous, and
each local likelihood is the linear regression model
\begin{equation} \label{eq:norm-i}
p(\x_i|\pai,\Thi,\hBs)=
  N(\x_i|m_i + \sum_{\x_j \in \pai} b_{ji} \x_j, 1/v_i)
\end{equation}
where $N(\x_i|\mu,\tau)$ is a normal distribution with mean $\mu$ and
precision $\tau>0$.  Given this form, a missing arc from $\X_j$ to
$\X_i$ implies that $b_{ji}=0$ in the complete DAG model.  The
local parameters are given by $\Thi=(m_i,\vecb_i,v_i)$, where
$\vecb_i$ is the column vector $(b_{1i}, \ldots, b_{i-1,i})$.

For Gaussian DAG models, the joint likelihood
$p(\C|\TBs,\hBs)$ obtained from Eqs~\ref{eq:bn-def-p}
and~\ref{eq:norm-i}
is an $n$-dimensional multivariate normal distribution
with mean $\vecmu$ and symmetric positive definite 
precision matrix $\matW$,
\[
p(\C|\TBs,\hBs) = \prod_{i=1}^n p(\x_i|\paSi,\Thi,\hBs)
= N(\C|\vecmu, \matW).
\]

For a complete model $\Bmc$ with ordering $(\X_1,\ldots,\X_n)$ 
there is a one-to-one mapping between $\TBsc = \bigcup_{i=1}^n \Thi$ 
where $\Thi=(m_i,\vecb_i,v_i)$ 
and $\{\vecmu,\matW\}$ which has a nowhere singular Jacobian matrix.
Consequently, assigning a prior for the parameters 
of one complete model induces a parameter prior, via the 
change of variables formula, for
$\{\vecmu,\matW\}$ and in turn, induces
a parameter prior for every complete model.
Any such induced parameter prior must satisfy,
according to our assumptions, global parameter independence.
Not many prior distributions satisfy such a requirement.  
In fact, in the next section we show
that the parameter prior $p(\vecmu,\matW|\hBsc)$ 
must be a normal-Wishart distribution.

For now we proceed by simply choosing $p(\vecmu,\matW|\hBsc)$
to be a normal-Wishart distribution.  In particular,
$p(\vecmu|\matW,\hBsc)$ is a multivariate-normal distribution
with mean $\nu$ and precision matrix $\amu
\matW$ ($\amu > 0$); and $p(\matW|\hBsc)$ is a Wishart
distribution, given by,
\begin{equation} \label{eq:cln} 
p(\matW|\hBsc)=
  c(n,\aw) |T|^{\aw/2} |\matW|^{(\aw-n-1)/2} e^{-1/2\tr\{T \matW\}}  
\end{equation}
with $\aw$ degrees of freedom $(\aw > n-1)$ and
a positive-definite precision matrix $T$
and where $c(n,\aw)$ is a normalization constant
given by
\begin{equation} 
\label{eq:normalization-constant}
c(n,\alpha) = 
\left[ 2^{\alpha n/2} \pi^{n(n-1)/4}
\prod_{i=1}^n \Gamma\left(\frac{\alpha+1-i}{2}\right)\right]^{-1}
\end{equation} 
(e.g., DeGroot, 1970, p. 57)\nocite{DeGroot70}.  

This choice satisfies global parameter independence
due to the following well known theorem.
Define a block partitioning $\{W_{11}, W_{12}, W'_{12}, W_{22}\}$ 
of an $n$ by $n$ matrix $W$ to be {\em compatible} with
a partitioning $\mu_1, \mu_2$ of an $n$ dimensional vector $\mu$,
if 
the indices of the rows that correspond to block $W_{11}$
are the same as the indices of the terms that constitute $\mu_1$
and similarly for $W_{22}$ and $\mu_2$.
\begin{theorem} \label{thm:pinw} 
If $f(\mu,W)$ is 
an $n$ dimensional normal-Wishart distribution, $n \geq 2$,
with parameters $\nu, \amu$, $\aw$, and $T$,
then $\{\mu_1, W_{11} - W_{12} W_{22}^{-1} W'_{12} \}$ 
is independent of 
$\{\mu_2 - W_{22}^{-1} W'_{12} \mu_1,$
$ W_{12},W_{22}\}$
for every partitioning $\mu_1, \mu_2$ of $\mu$ where
$W_{11}$,$W_{12}$, $W'_{12}$, $W_{22}$ is 
a block partitioning of $W$
compatible with the partitioning $\mu_1,\mu_2$.  
Furthermore, the pdf of
$\{\mu_1, W_{11} - W_{12} W_{22}^{-1} W'_{12} \}$ 
is normal-Wishart with parameters
$\nu_1$, $\amu$, $T_{11} - T_{12} T_{22}^{-1} T'_{12} $,
and $\aw -n +l$ where 
$T_{11}$,$T_{12}$, $T'_{12}$, $T_{22}$ is 
a compatible block partitioning of $T$,
$\nu_1, \nu_2$ is a compatible partitioning of $\nu$, and
$l$ is the size of the vector $\nu_1$.
\end{theorem}

The proof of Theorem~\ref{thm:pinw} 
requires a change of variables from
$(\mu,W)$ to $(\mu_1$, $\mu_2 - W_{22}^{-1} W'_{12} \mu_1)$
and $(W_{11} - W_{12} W_{22}^{-1} W'_{12},$ $W_{12},W_{22})$.
Press carries out these computations for the Wishart distribution
(1971, p. 117-119)\nocite{Press71}. 
Standard changes are needed
to obtain the claim for the normal-Wishart distribution.

To see why the independence conditions
in Theorem~\ref{thm:pinw} imply global parameter
independence, consider the partitioning in which 
the first block contains the first $n-1$
coordinates which correspond to $X_1,\ldots, X_{n-1}$ 
while the second block contains the last
coordinate which corresponds to $X_n$.
For this partitioning, $b_n = - W_{22}^{-1} W'_{12}$,
$v_n =  W_{22}^{-1}$, and $m_n = \mu_2 - W_{22}^{-1} W'_{12} \mu_1$.
Furthermore, 
$((W^{-1})_{11})^{-1} = W_{11} - W_{12} W_{22}^{-1} W'_{12}$ is
the precision matrix associated with $X_1,\ldots, X_{n-1}$.
Consequently, $\{m_n, b_n,v_n\}$
is independent of $\{\mu_1,((W^{-1})_{11})^{-1}\}$.
We now recursively repeat this argument with
$\{\mu_1, ((W^{-1})_{11})^{-1}\}$ instead of $\{\mu, W\}$,
to obtain global parameter independence.
The converse, namely that global parameter independence 
implies the independence conditions in Theorem~\ref{thm:pinw},
is established similarly.

\comment{ Old proof of parameter independence

\begin{theorem} \label{thm:pinw} 
If $(\vecmu,\matW)$ has a normal--Wishart probability distribution, then
$p(\vecm, \vecv, \matB)$$ = \prod_{i=1}^n p(m_i, v_i, \vecb_i)$
where 
$\vecm = \{ m_i, 1 \leq i \leq n \}$,
$\vecv = \{ v_i, 1 \leq i \leq n \}$, and
$\matB = \{ \vecb_i, 1 \leq i \leq n \}$.
\end{theorem} 
 
\noindent {\bf Proof Sketch:} 
Compute
$p(\vecm|\vecv,\matB)$ and $p(\vecv,\matB)$ from
$p(\vecmu,\matW)$ using a change a variable formula.
The one-to-one mapping between
$\{\vecm,\vecv,\matB\}$ and 
$\{\vecmu,\matW\}$ is given by
$\mu_i = m_i + \sum_{j=1}^{i-1} b_{ji} \mu_j$
and the recursive formula,
$\matW(1)  =  1/v_1$,
\[
\matW(i+1)  =  \left( \begin{array}{cc} 
\matW(i) + \frac{\vecb_{i+1} \vecb^t_{i+1}}{v_{i+1}} & 
-\frac{\vecb_{i+1}}{v_{i+1}}   \\ 
-\frac{\vecb^t_{i+1}}{v_{i+1}} &  \frac{1}{v_{i+1}} 
\end{array} \right), \ \ \ i>1 \nonumber
\]
where $W(i)$ is the $i \times i$ upper left submatrix of $\matW$
(e.g., Shachter and Kenley, 1989\nocite{Shachter89b}).
Press essentially carries out
these computations for the Wishart distribution
(1971, p. 117-119)\nocite{Press71}.  \qed
}  

Our choice of prior implies that
the posterior $p(\vecmu,\matW|\Db,\hBsc)$ is also a normal-Wishart
distribution (DeGroot, 1970, p. 178)\nocite{DeGroot70}.  
In particular, 
$p(\vecmu|\matW,\Db,\hBsc)$ is multivariate normal with mean vector
given by
\begin{equation} 
\label{eq:updatemeans} 
\frac{\amu \nu + m \Xvecbar_m}{\amu+m}
\end{equation} 
and precision matrix $(\amu+m)\matW$, where $\Xvecbar_m$ is the sample
mean of $\Db$, and $p(\matW|\Db,\hBsc)$ is a Wishart distribution with
$\aw+m$ degrees of freedom and precision matrix $R$ given by
\begin{equation} \label{eq:updateTau} 
R =  
T + \matS_m + \frac{\amu m}{\amu+m}
  (\nu -\Xvecbar_m)(\nu -\Xvecbar_m)' 
\end{equation} 
where 
$\matS_m= \sum_{\textcolor{blue}{l}=1}^m (\Xvecsubl -\Xvecbar_m)(\Xvecsubl -\Xvecbar_m)'$.
From these equations, we see
that $\amu$ and $\aw$ can be thought 
of as \textcolor{blue}{effective sample sizes for
the normal and Wishart components of the prior, respectively.}

\comment{
Given $\bY \subseteq \U$ ($|\bY|=l$), and vector ${\bf
z}=(z_1,\ldots,z_n)$, let ${\bf z}_{\bY}$ denote the vector formed by
the components $z_i$ of ${\bf z}$ such that $\X_i \in \bY$.
Similarly, given matrix $\matM$, let $\matM_{\bY \bY}$ denote the
submatrix of $\matM$ containing elements $m_{ij}$ such that $\X_i,\X_j
\in \bY$.  If 
$p(\vecmu,\matW|\hBsc)$ is a normal-Wishart distribution as we have
described, then $p(\vecmu_{\bY},((\matW^{-1})_{\bY \bY})^{-1}|\hBsc)$ is
also a normal--Wishart distribution with 
hyper parameters $\nu_{\bY}$, 
$\amu$, $T_{\bY}=((T^{-1})_{\bY \bY})^{-1}$ and $\awY=\aw-n+l$
(e.g., see Press, 1971, Theorems 5.1.3 and 5.1.4).  
}

According to Theorem~\ref{thm:pinw},
if $p(\vecmu,\matW|\hBsc)$ is a 
normal-Wishart distribution with the parameters
given by the theorem, 
then $p(\vecmu_{\bY},((\matW^{-1})_{\bY \bY})^{-1}|\hBsc)$ is
also a normal--Wishart distribution with
effective same sizes $\amu$ and $\aw-n+l$, and 
parameters $\nu_{\bY}$ and
$\textcolor{red}{T_{\bY \bY}}$, 
where $\bY$ is a subset of $l$ coordinates \textcolor{blue}{and $M_{\bY \bY}$ is the matrix $M$
with elements restricted to the corresponding variables $\bY$.}
Thus, we obtain the terms in \Eq{eq:be}:
\begin{equation} \label{eq:bge}
p(\Db^{\bY}|\hBsc)  = (2\pi)^{-lm/2} \ \left(\frac{\amu}{\amu+m}\right)^{l/2} \ 
  \frac{c(l,\aw-n+l)}{c(l,\aw-n+l+m)} \ 
  |\textcolor{red}{T_{\bY \bY}}|^{\frac{\aw-n+l}{2}} \ 
  |\textcolor{red}{R_{\bY \bY}}|^{-\frac{\aw-n+l+m}{2}}
  \nonumber
\end{equation}
\textcolor{blue}{(See Geiger and Heckerman, 1994, for a derivation
when $l=n$.)}

\textcolor{blue}{For the assessment of the parmeter prior,  we consider
a partially indirect approach.  
We start with the observation that when
$p(\vecmu,\matW|\hBsc)$ is normal--Wishart as we have described, then
then $p(\C|\hBsc)$ is a multivariate $t$ distribution with $\aw-n+1$
degrees of freedom, location vector $\nu$, and precision matrix
$\amu(\aw-n+1)/(\amu+1)T^{-1}$. 
This result can be derived by first integrating over $\vecmu$ using
Equation 6 on p.\ 178 of DeGroot with sample size equal to one, and then
integrating over $\matW$
following an approach similar to that on pp.\ 179--180. Next, when
$\aw > n+1$, it follows that
\begin{equation} \label{eq:t1}
{\rm E}(\C|\hBsc) = \nu \ \ \ \ \ \ 
{\rm Cov}(\C|\hBsc) = \frac{\amu+1}{\amu} \ \frac{1}{\aw-n-1} \ T
\end{equation}
(e.g., DeGroot, 1970, pp.\ 61).  
Thus, a person can assess the parameter prior
by assessing $\amu$ and $\aw$, driectly, and by assessing 
a DAG model for E$(\C|\hBsc)$ and Cov$(\C|\hBsc)$ and
then computing $\nu$ and $T$ using Equations~\ref{eq:t1}.
We call this model a {\em prior} DAG model.
The unusual aspect of this assessment is the conditioning hypothesis
$\hBsc$ (see Heckerman et al. [1995a] for a discussion).
This indirect approach provides a suitable Bayesian
alternative for many of the examples discussed in Spirtes, Glymour,
and Scheines (1993).\nocite{Spirtes93}}

\section{Characterization of Several Probability Distributions}


We now characterize the Wishart distribution 
as the only pdf that satisfies global parameter independence
for an unknown precision matrix $W$ with $n \geq 3$ coordinates
(Theorem~\ref{thm:Wishart}).
This theorem is phrased and proven in a
terminology that relates to known facts about the
Wishart distribution.
We proceed with similar characterizations of the
normal and normal-Wishart distributions
(Theorems~\ref{thm:normal} and~\ref{thm:normalWishart}).

\begin{theorem} \label{thm:Wishart}
Let $W$ be an $n \times n$, $n \ge 3$, positive-definite symmetric 
matrix of random variables
and $f(W)$ be a pdf of $W$.
Then, f$(W)$ is a Wishart distribution if and only if
$W_{11} - W_{12} W_{22}^{-1} W'_{12}$ is independent of 
$\{W_{12},W_{22}\}$
for every block partitioning $W_{11},W_{12}, W'_{12}, W_{22}$ of
$W$.
\end{theorem} 

\noindent {\bf Proof:} 
That
$W_{11} - W_{12} W_{22}^{-1} W'_{12}$ is independent 
of $\{W_{12},W_{22}\}$
whenever $f(W)$ is a Wishart distribution is a well
known fact (Press 1971, p. 117-119)\nocite{Press71}. 
It is also expressed by Theorem~\ref{thm:pinw}.
The other direction is proven by induction on $n$.
The base case $n=3$ is treated at the end.

The pdf of $W$ can be written in $n!$ orderings.
In particular, due to the assumed independence conditions,
we have the following equality:
{\small
\begin{eqnarray}
\label{eq:wish1}
\lefteqn{
\nonumber
f(W) = f_1(\A - \B \D^{-1} \B') f_{2|1}(\D,\B)
}
\\
& & 
= f_2(\D - \B' \A^{-1} \B) f_{1|2}(\A,\B) 
\end{eqnarray}
}
where a subscripted $f$ denotes a pdf.
Since $n > 3$, we can divide the indices of $W$
into three non-empty sets $a,b$ and $c$ such that $b$
includes at least two indices.  We now
group $a$ and $b$ to form a block and $b$ and $c$ to form a block.
For each of the two cases, let $W_{11}$ be the block 
consisting of the indices in $\{a,b\}$ or $\{b,c\}$, respectively,
and $W_{22}$ be the block consisting
of the indices of $c$ or $a$, respectively.
By the induction hypothesis, and since the independence conditions
on $W$ can be shown to hold for any block $W_{11}$ of $W$,
we conclude that
$f_1(V)$ is a Wishart distribution
$W(V |\; \alpha_1,T_{1})$ and $f_2(V)$ is a Wishart distribution
$W(V |\; \alpha_2,T_{2})$.  Consequently, the pdf of 
the block corresponding to the indices in $b$ is
a Wishart distribution, and from the two alternative ways
by which this pdf can be formed, 
it follows that $\alpha_1 - l_1 = \alpha_2 - l_2$,
where $l_i$ is the number of indices in block $i$
(Press, 1971, Theorem 5.1.4).
Thus,
\begin{eqnarray}
\label{eq:wish2}
\nonumber
\lefteqn{
c_1 |W_{11.2}|^{\beta}
e^{\tr\{T_{1} W_{11.2}\}} f_{2|1}(\D,\B)  = }
\\
\;\;\;
& & 
c_2 |W_{22.1}|^{\beta} 
e^{\tr\{T_{2} W_{22.1}\}} f_{1|2}(\A,\B) 
\end{eqnarray}
where $c_1$ and $c_2$ are normalizing constants,
$\beta= (\alpha_1 - l_1 -1)/2$,
$W_{11.2} = \A-\B\D^{-1}\B'$,
and
$W_{22.1}= \D - \B'\A^{-1} \B $.
Define
{\small
\begin{eqnarray}
\lefteqn{
\label{eq:wish3}
F_{2|1}(\D,\B) =
}
\\
\nonumber
& & c_1 f_{2|1}(\D,\B) / |\D|^{\beta}
e^{\tr\{T_{2} \D + T_{1} (\B\D^{-1}\B')\}}
\end{eqnarray}
\begin{eqnarray}
\lefteqn{
\label{eq:wish3.1}
F_{1|2}(\A,\B) = 
}
\\
\nonumber
& &  c_2 f_{1|2}(\A,\B) / |\A|^{\beta}
e^{\tr\{T_{1} \A + T_{2} (\B' \A^{-1} \B) \}},
\end{eqnarray}
} 
substitute into Equation~\ref{eq:wish2}, and obtain,
using $|\A-\B\D^{-1}\B'| |\D| = |W|$, that
$F_{2|1}(\D,\B) = F_{1|2}(\A,\B)$.
\comment{
\begin{equation}
\label{eq:wish3.5}
F_{2|1}(\D,\B) = F_{1|2}(\A,\B)
\end{equation}
}
Consequently, $F_{2|1}$ and $F_{1|2}$
are functions only of $\B$ and thus, using
Equation~\ref{eq:wish3}, we obtain
\begin{equation}
\label{eq:wish4}
f(W) = 
|W|^{\beta} e^{\tr\{T_{1} \A + T_{2} \D\}} H(\B)
\end{equation}
for some function $H$.
\comment{
, or equivalently,
\begin{equation}
\label{eq:wish5}
f(W) = 
c |W|^{\beta} e^{\tr\{T W\}} h(\B)
\end{equation}
where $h(\B)= e^{\tr\{ T_{12} \B' + T'_{12} \B\}} H(\B)$.
}

To show that $f(W)$ is Wishart we must
find the form of $H$.  
Considering the three possible pairs of blocks 
formed with the sets of indices $a$, $b$, and $c$,
Equation~\ref{eq:wish4} can be rewritten as follows.
\begin{eqnarray}
\lefteqn{
\label{eq:wish6}
f(W) = |W|^{\beta_1} e^{\tr 
\{T_{aa} W_{aa} + T_{bb} W_{bb} + T_{cc} W_{cc} \}} 
\cdot
}
\\
\nonumber
& &
e^{2 \tr \{T'_{ab} W_{ab} + T'_{ac} W_{ac} + T'_{bc} W_{bc} \}} 
H_1(W_{ac}, W_{bc})
\end{eqnarray}
\begin{eqnarray}
\lefteqn{
\label{eq:wish7}
f(W) = 
|W|^{\beta_2} e^{ \tr
\{S_{aa} W_{aa} + S_{bb} W_{bb} + S_{cc} W_{cc} \}} 
\cdot
}
\\
\nonumber
& &
e^{2 \tr \{S'_{ab} W_{ab} + S'_{ac} W_{ac} + S'_{bc} W_{bc} \}} 
H_2(W_{ab}, W_{bc})
\end{eqnarray}
\begin{eqnarray}
\lefteqn{
\label{eq:wish8}
f(W) = 
|W|^{\beta_3} e^{ \tr
\{R_{aa} W_{aa} + R_{bb} R_{bb} + T_{cc} W_{cc} \}} 
\cdot
}
\\
\nonumber
& &
e^{2 \tr \{R'_{ab} W_{ab} + R'_{ac} W_{ac} + R'_{bc} W_{bc} \}}
H_3(W_{ab}, W_{ac})
\end{eqnarray}
By setting 
$W_{ab}= W_{ac} = W_{bc}=0$, we get
$\beta_1= \beta_2 = \beta_3$ and
$T_{ii} = S_{ii} = R_{ii}$, for $i=a,b,c$.
By comparing Equations~\ref{eq:wish6} and~\ref{eq:wish7}
we obtain
\begin{eqnarray}
\lefteqn{
\label{eq:wish9}
e^{2 \tr \{ (T'_{ac} - S'_{ac}) W_{ac}\}} 
H_1(W_{ac}, W_{bc})
=
}
\\
\nonumber
& &
e^{2 \tr \{
(S'_{ab} - T'_{ab}) W_{ab} + 
(S'_{bc} - T'_{bc}) W_{bc} \}} 
H_2(W_{ab}, W_{bc})
\end{eqnarray}
Each side of this equation must be a function
only of $W_{bc}$. We denote this function by $H_{12}$.
Hence,
\[
H_1(W_{ac}, W_{bc}) = 
H_{12}(W_{bc})
e^{2 \tr \{ (S'_{ac} - T'_{ac}) W_{ac}\}} 
\]
and by symmetric arguments,
comparing Equations~\ref{eq:wish6} and~\ref{eq:wish8},
\[
H_1(W_{ac}, W_{bc}) = 
H_{13}(W_{ac})
e^{2 \tr \{ (R'_{bc} - T'_{bc}) W_{bc}\}} 
\]
Thus,
$H_{12}(W_{bc})$ is proportional 
to  $e^{2 \tr \{ (R'_{bc} - T'_{bc}) W_{bc}\}}$
and so $f(W)$ is a Wishart distribution, as claimed.

It remains to examine the case $n=3$.  
We first assume $n=2$ in which case
$f(W)$ is not necessarily a Wishart distribution.
In the full version of this paper (Submitted to Annals
of Statistics)
we show that given the independence conditions for
two coordinates, $f$ must have the form
\begin{equation}
\label{eq:basis1}
f(W) = c |W|^{\beta} e^{\tr\{T W\}} H(\B)
\end{equation}
where $H$ is an arbitrary function,
and that the marginal distributions of
$W_{11}$ and $W_{22}$
are one dimensional Wishart distributions.
The proof rests on techniques from the theory of 
functional equations 
(\Aczel, 1966) \nocite{Ac66} 
and results from 
(\Jarai, 1986, 1998). \nocite{Ja86,Ja98}
A weaker proof, under some regularity conditions,
can be found in (Geiger and Heckerman, 1998).
\nocite{GH98pms}

We now treat the case $n=3$ using 
these assertions about the case $n=2$.
Starting with Equation~\ref{eq:wish1},
and proceeding with blocks $a,b,c$ each containing exactly 
one coordinate,
we get, due to the given independence conditions for
two coordinates, that $f_1$ has the form
given by Equation~\ref{eq:basis1}, and that
$f_2$ is a one dimensional Wishart distribution.
Proceeding parallel to Equations~\ref{eq:wish2}
through~\ref{eq:wish3.1}, we obtain,

\begin{equation}
\label{eq:basis2}
H(a_{12} - b^2_1 b^2_2/W_{22}) F_{2|1}(\D,\B)
= F_{1|2}(\A,\B)
\end{equation}
where
$(b_1,b_2)$ is the matrix $W_{12}$, 
$a_{12}$ is the off-diagonal element of $W_{11}$,
$a_{12} - b^2_1 b^2_2/W_{22}$ is the off diagonal element
of $W_{11}- W_{12} W^{-1}_{22} W'_{12}$,
and $W_{22}$ is a $1\times 1$ matrix.
Note that the right hand side depends on $W_{11}$ only
through $a_{12}$.  Let $b_1$ and $b_2$ be 
fixed,
\comment{
that $F_{1|2}(\A,(b_1,b_2))$ is not identically zero,
Alternative 1: 
let $y= -b^2_1 b^2_2/W_{22}$, and $x= a_{12}$.  Also let
$F(t)= F_{2|1}(-b^2_1 b^2_2/t,(b_1 ,b_2))$
and
$G(a_{12}) = F_{1|2}(\A,(b_1,b_2))$.
We can now rewrite Equation~\ref{eq:basis2} as
$H(x+y) F(y) = G(x)$.  
Since $F$ is positive everywhere, 
we can define $K(y) = 1 /F(y)$,
and rewrite the last equation as
\begin{equation}
\label{eq:basis5}
H(x+y) = G(x) K(y)
\end{equation}
the only measurable solution of which for $H$
is $H(z) = c e^{bz}$ (e.g., \Aczel, 1966) \nocite{Ac66}.

Alternative 2: 
}
$y= b^2_1 b^2_2/W_{22}$, and $x= a_{12}$.  Also let
$F(t)= F_{2|1}(b^2_1 b^2_2/t,(b_1 ,b_2))$
and
$G(a_{12}) = F_{1|2}(\A,(b_1,b_2))$.
We can now rewrite Equation~\ref{eq:basis2} as
$H(x-y) F(y) = G(x)$.  
Now set $z=x-y$, and obtain
for every $y,z >0$
\begin{equation}
\label{eq:basis5}
H(z) F(y) = G(y+z)
\end{equation}
the only measurable solution of which for $H$
is $H(z) = c e^{bz}$ (e.g., \Aczel, 1966) \nocite{Ac66}.

Substituting this form of $H$
into Equation~\ref{eq:basis1},
we see that $W_{11}$ has a two
dimensional Wishart distribution.
Recall that $W_{22}$ has a one dimensional
Wishart distribution.  We can now apply the
induction step starting form Equation~\ref{eq:wish2}
and prove the Theorem for $n=3$.  \qed

We now treat the situation when only the means are unknown,
characterizing the normal distribution.
The two dimensional case turns out to be
covered by the
Skitovich-Darmois theorem 
(e.g., Kagan, Linnik, and Rao (1973)).
\nocite{KLR73}

\begin{theorem}[Skitovich-Darmois] \label{thm:Skitovich-Darmois}
Let $z_1,\ldots,z_k$ be independent random variables and
$\alpha_i, \beta_i$, $1<i<k$, be constant coefficients.
If $L_1 = \sum \alpha_i z_i$
is independent of $L_2 = \sum \beta_i z_i$,
then each $z_i$ for which $\alpha_i \beta_i \neq 0$
is normal.
\end{theorem}

The Skitovich-Darmois theorem is used in the proof of
the base case of our next characterization.
Several generalizations of the Skitovich-Darmois theorem
are described in Kagan et al.\ (1973).  \nocite{KLR73}

\begin{theorem} \label{thm:normal}
Let $W$ be an $n \times n$, $n \ge 2$, positive-definite symmetric 
matrix of real random variables such that no entry in $W$ is zero,
$\mu$ be an $n$-dimensional vector of random variables,
and $f(\mu)$ be a pdf of $\mu$.
Then, f$(\mu)$ is an $n$ dimensional normal distribution 
$N(\mu | \eta, \gamma W)$ where $\gamma >0$ if and only if
$\mu_1$ is independent of $\mu_2 + W_{22}^{-1} W'_{12} \mu_1$
for every partitioning $\mu_1, \mu_2$ of $\mu$ where
$W_{11}$,$W_{12}$, $W'_{12}$, $W_{22}$ is a 
block partitioning of $W$ compatible
with the partitioning $\mu_1, \mu_2$.

\end{theorem} 


\noindent {\bf Proof:} 
The two independence conditions,
$\mu_1$ independent of $\mu_2 + W_{22}^{-1} W'_{12} \mu_1$
and 
$\mu_2$ independent of $\mu_1 + W_{11}^{-1} W_{12} \mu_2$,
are equivalent to the following functional equation
\begin{eqnarray}
\lefteqn{
\label{eq:normal1}
f(\mu) =
f_1(\mu_1) f_{2|1}( \mu_2 + W_{22}^{-1} W'_{12} \mu_1)
}
\\
\nonumber
& &
= f_2(\mu_2) f_{1|2}( \mu_1 + W_{11}^{-1} W_{12} \mu_2)
\end{eqnarray}
where a subscripted $f$ denotes a pdf.
We show that the only solution for $f$ that satisfies
this equation is the normal distribution.
Consequently both the if and only if portions
of the theorem will be established.

For $n \geq 3$, we can divide the indices of $W$
into three non-empty sets $a,b$ and $c$.
We group $a$ and $b$ to form a block and $b$ and $c$ to form a block.
For each of the two cases, let $W_{11}$ be the block 
consisting of the indices in $\{a,b\}$ or $\{b,c\}$, respectively,
and $W_{22}$ be the block consisting
of the indices of $c$ or $a$, respectively.
By the induction hypothesis applied to both cases
and marginalization we can assume that
$f_1(\mu_1)$ is a normal distribution
$N(\mu_1 | \eta_1, \gamma_1 (W^{-1})_{11})^{-1})$ 
and that
$f_2(\mu_2)= 
N(\mu_2 | \eta_2, \gamma_2 (W^{-1})_{22})^{-1})$.
Consequently, the pdf of 
the block corresponding to the indices in $b$ is
a normal distribution, and from the two alternative ways
by which this pdf can be formed, 
it follows that $\gamma_1 = \gamma_2$.

Let $\gamma = \gamma_i$, $i=1,2$, and define
\begin{eqnarray}
\nonumber
\label{eq:normal3}
F_{2|1}( x ) & = & 
f_{2|1}( x ) / 
N( x | \eta_2 +  W_{22}^{-1} W'_{12} \eta_1,  \gamma W_{22})
\\
\nonumber
\label{eq:normal4}
F_{1|2}( x ) & = & 
f_{1|2}( x ) /
N( x | \eta_1 + W_{11}^{-1} W_{12} \eta_2,  \gamma W_{11}).
\end{eqnarray}
By substituting these definitions into Equation~\ref{eq:normal1},
substituting the normal form for 
$f_1(\mu_1)$ and $f_2(\mu_2)$, 
and canceling 
on both sides of the equation 
the term $N(\mu|\eta,\gamma W)$ 
(which is formed by standard algebra pertaining 
to quadratic forms (E.g., DeGroot, pp. 55)),
we obtain a new functional equation, 
\[
F_{2|1}( \mu_2 + W_{22}^{-1} W'_{12} \mu_1)
=
F_{1|2}( \mu_1 + W_{11}^{-1} W_{12} \mu_2).
\]
By setting $\mu_2 = - W_{22}^{-1} W'_{12} \mu_1$, we 
obtain
$F_{1|2}( (I - (W_{11}^{-1} W_{12})(W_{22}^{-1} W'_{12})) \mu_1)=
F_{2|1}( 0 )$ for every $\mu_1$.
Hence, the only solution to this functional equation is
$F_{1|2} =F_{2|1} \equiv \mbox{constant}$.
Consequently, $f(\mu) = N(\mu|\eta, \gamma W)$.

It remains to prove the theorem for $n=2$.
Let 
$z_1 = \mu_1$,
$z_2 = \mu_2 + w_{22}^{-1} w_{12} \mu_1$,
$L_1 = \mu_1 + w_{11}^{-1} w_{12} \mu_2$,
and 
$L_2 = \mu_2$.
By our assumptions $z_1$ and $z_2$ are independent
and $L_1$ and $L_2$ are independent.
Furthermore, rewriting $L_1$ and $L_2$ in terms of $z_1$ and $z_2$,
we get,
$L_1 = w^{-1}_{11} w^{-1}_{22} (w_{11} w_{22} - w^2_{12}) z_1 
+ w_{11}^{-1} w_{12} z_2$
and
$L_2 = z_2 -  w_{22}^{-1} w_{12} z_1$.
All linear coefficients in this transformation
are non zero due to the fact that $W$
is positive definite and that $w_{12}$ is not zero.
Consequently, due to the Skitovich-Darmois theorem,
$z_1$ is normal and $z_2$ is normal. 
Furthermore, since $z_1$ and $z_2$ are independent,
their joint pdf is normal as well.
Finally, $\{\mu_1, \mu_2\}$ and $\{z_1,z_2\}$ are
related through a non-singular linear transformation
and so $\{\mu_1,\mu_2\}$ also have a joint normal distribution
$f(\mu) = N(\mu|\eta,A)$
where $A= (a_{ij})$ is a $2 \times 2$ precision matrix.
Substituting this solution into
Equation~\ref{eq:normal1} and comparing the coefficients of 
$\mu_1^2$, $\mu_2^2$, and $\mu_1 \mu_2$, we obtain
$a_{12}/a_{11} = w_{12}/w_{11}$ and $a_{12}/a_{22} = w_{12}/w_{22}$.
Thus $A= \gamma W$ where $\gamma >0$.
\qed

The proofs of Theorems~\ref{thm:Wishart} and~\ref{thm:normal}
can be combined to form the following characterization
of the normal-Wishart distribution.

\begin{theorem} \label{thm:normalWishart}
Let $W$ be an $n \times n$, $n \ge 3$, positive-definite symmetric 
matrix of real random variables such that no entry in $W$ is zero,
$\mu$ be an $n$-dimensional vector of random variables,
and $f(\mu, W)$ be a joint pdf of $\{\mu, W\}$.
Then, $f(\mu,W)$ is 
an $n$ dimensional normal-Wishart distribution 
if and only if
$\{\mu_1, W_{11} - W_{12} W_{22}^{-1} W'_{12} \}$ 
is independent of 
$\{\mu_2 + W_{22}^{-1} W'_{12} \mu_1, W_{12},W_{22}\}$
for every partitioning $\mu_1, \mu_2$ of $\mu$ where
$W_{11}$,$W_{12}$, $W'_{12}$, $W_{22}$ is a
block partitioning of $W$ compatible
the partitioning $\mu_1, \mu_2$.
\end{theorem} 

\noindent {\bf Proof:} 
The two independence conditions,
$\{\mu_1, W_{11} - W_{12} W_{22}^{-1} W'_{12} \}$ 
independent of 
$\{\mu_2 + W_{22}^{-1} W'_{12} \mu_1, $ $W_{12},W_{22}\}$
and 
$\{\mu_2, W_{22} - W'_{12} W_{11}^{-1} W_{12} \}$ 
independent of 
$\{\mu_1 + W_{11}^{-1} W_{12} \mu_2,$ $W'_{12},W_{11}\}$,
are equivalent to the following functional equation
{\small
\begin{eqnarray}
\label{eq:nw1}
\lefteqn{
\nonumber
f(\mu,W) = f_1(\mu_1, W_{11.2}) 
f_{2|1}(\mu_2 + \D^{-1} \B' \mu_1,\D,\B) 
}
\\
\nonumber
& & 
\;\;\;\;\;\;
= f_2(\mu_2,W_{22.1}) 
f_{1|2}(\mu_1 + \A^{-1} \B \mu_2,\A,\B)
\end{eqnarray}
}
where a subscripted $f$ denotes a pdf.
We show that the only solution for $f$ that satisfies
this functional equation is the normal-Wishart distribution.
Setting $W$ to a fixed value yields
Equation~\ref{eq:normal1}
the solution of which for $f$ is 
proportional to $N(\mu | \eta, \gamma W)$.
Similarly, the solutions for the functions
$f_1, f_2, f_{1|2} $, and $f_{2|1}$ are
also proportional to normal pdfs.
The constants $\eta$ and $\gamma$ could potentially
change from one value of $W$ to another.
However, since $\eta_1$ can only be a function
of $\A - \B \D^{-1} \B'$ due to the solution for $f_1$,
and since it must also be a function of 
$\{\D,\B\}$ due to the solution for $f_{2|1}$,
it cannot change with $W$. Similarly $\eta_2$
cannot change with $W$.
Substituting this solution into
Equation~\ref{eq:nw1}
and dividing by the common terms
which are equal to $f(\mu | W)$ yields
Equation~\ref{eq:wish1}
the solution of which for $f$ is a Wishart pdf.
\qed

Note that the conditions set on $W$
in Theorem~\ref{thm:normalWishart}, namely,
a positive-definite symmetric 
matrix of real random variables such that no entry in $W$ is zero,
are necessary and sufficient in order for $W$ to be 
a precision matrix of a complete Gaussian DAG model.

\section{Local versus Global Parameter Independence}

We have shown that the only pdf for $\{\mu,W\}$  
which satisfies global parameter independence, when the number
of coordinates is greater than two, is the normal-Wishart
distribution. We now discuss additional independence assertions
implied by the assumption of 
global parameter independence.

\comment{ Old definition -- more technical than needed
\vspace{1ex}
\noindent
{\bf Definition}
{\em Local parameter independence}
is the assertion that for every DAG model $m$ for $X_1,\ldots, X_n$,
and for every $i=1,\ldots, n$, there exists an $l_i>1$ such that
$f_i(\theta_{i1},\ldots,\theta_{il_i})= \prod_{j=1}^{l_i} 
f_{ij}(\theta_{ij})$,
where $\{\theta_{ij}, 1 \leq j \leq l_i\}$
is a set of parameters defining the i-th local distribution
for $m$ and $f_i$ is their joint prior distribution.

\vspace{1ex}

In other words, this definition asserts that the prior
of the parameters of each local probability distribution
factors to at least two terms.
}

\vspace{1ex}
\noindent
{\bf Definition}
{\em Local parameter independence}
is the assertion that for every DAG model $m$ for $X_1,\ldots, X_n$,
there exists a partition of the parameters of each local distribution
into at least two independent sets.

\vspace{1ex}

Consider the parameter prior for
$\{m_n, b_n,v_n\}$ when the prior for $\{\mu,W\}$
is a normal Wishart as specified by 
Equations~\ref{eq:cln} and~\ref{eq:normalization-constant}.
By a change of variables, we get
{\small
\begin{eqnarray*}
\lefteqn{
f_n(m_n, b_n,v_n) =
W(1/v_n \;|\; \alpha+n-1, T_{22} - T'_{12} T_{11}^{-1} T_{12}) 
\cdot 
}
\\
\nonumber
& & 
\;\;\;\;\;\;
N(b_n \;|\; T_{11}^{-1} T_{12}, T_{22}/v_n)
\cdot 
N(m_n \;|\; \nu_n, \amu/v_n)
\end{eqnarray*}
}
where the first block corresponds to $X_1,\ldots,X_{n-1}$
and the second block corresponds to $X_n$.
We note that the only independence assumption
expressed by this product is that $m_n$ and $b_n$ 
are independent given $v_n$.  However, by standardizing
$m_n$ and $b_n$, namely defining, 
$m^*_n = (m_n - \nu_n )/(\amu/v_n)^{1/2}$ 
and $b^*_n =   (T_{22}/v_n)^{1/2} (b_n -  T_{11}^{-1} T_{12})$,
which is well defined because $T_{22}$ is positive definite 
and $v_n >0$,
we obtain a set of parameters $(m^*_n, b^*_n, v_n)$
which are mutually independent. 
Furthermore, this mutual independence property
holds for every local family
and for every Gaussian DAG model over $X_1,\ldots, X_n$.
We call this property
the {\em standard local independence} for Gaussian DAG models.

This observation leads to the following corollary
of our characterization theorems.

\begin{corollary}
If global parameter independence holds for every 
complete Gaussian DAG model
over $X_1,\ldots,X_n$ ($n \geq 3)$, then standard local
parameter independence also holds for every complete 
Gaussian DAG model over $X_1,\ldots,X_n$.
\end{corollary}

This corollary follows from the fact that global parameter independence 
implies that, due to Theorem~\ref{thm:normalWishart}, 
the parameter prior is a normal-Wishart, and
for this prior, we have shown that
standard local parameter independence must hold.

It is interesting to note that when $n=2$, there are distributions
that satisfy global parameter independence but do not satisfy 
standard local parameter independence. In particular, a prior
for a $2 \times 2$ positive definite matrix $W$ 
which has the form $W(W | \alpha, T) H(w_{12})$,
where $H$ is some real function and $w_{12}$ is 
the off-diagonal element of $W$,
satisfies global parameter independence
but need not satisfy standard local parameter independence.
Furthermore, if standard local parameter independence is assumed, then
$H(w_{12})$ must be proportional to $e^{a w_{12}}$,
which means that, for $n=2$, the only pdf
for $W$ that satisfies global and standard 
local parameter independence
is the bivariate Wishart distribution.
In contrast, for $n>2$, global parameter
independence alone implies a Wishart prior.

\section{Discussion}

The formula for the marginal likelihood
applies whenever 
Assumptions~\ref{ass:cme} through~\ref{ass:pi}
are satisfied, not only for Gaussian DAG models.
Another important special case is when all variables in $\U$
are discrete and  all local distributions are
multinomial.  This case has been treated in
(Heckerman et al.\ (1995; Geiger and Heckerman, 1997)
\nocite{HGC95ml,GH97stat} 
under the additional assumption of
local parameter independence.
Our generalized derivation herein dispenses this assumption
and unifies the derivation in the discrete case with 
the derivation needed for Gaussian DAG models.

Furthermore, our proof also suggests
that the only parameter prior
for complete discrete DAG models with $n \geq 3$ variables
that satisfies Assumptions~\ref{ass:cme} through~\ref{ass:pi}
is the Dirichlet distribution.  
The added assumption of
local parameter independence, which 
is essential
for the characterization of the Dirichlet distribution
when $n=2$ (Geiger and Heckerman, 1997),
\nocite{GH97stat} 
seems to be redundant when $n \geq 3$,
just as it is redundant
for the characterization of the normal-Wishart distribution.

Our characterization means that
the assumption of global parameter independence 
when combined with the definition of
$\hBs$, the assumption of complete model equivalence,
and the regularity assumption, may be too restrictive.
One common remedy for this problem is to 
use a hierarchical prior $p(\theta| \eta) p(\eta)$
with hyperparameters $\eta$.
When such a prior is used for Gaussian DAG models
our results show that
for every value of $\eta$ for which
global parameter independence holds,
$p(\theta | \eta )$ must be a normal-Wishart distribution.
Another possible approach is to select one representative DAG
model from each class of equivalent DAG models, assume global
parametr independence only for these representatives,
and evaluate the marginal likelihood only
for these representatives. 
The difficulty with this approach is that
when projecting a prior from a complete DAG model
to a DAG model with missing edges, one needs to perform
additional high dimensional integrations, before using the parameter
modularity property (see Section~2). The assumption
of global parameter independence for all complete DAGs
rather than one, removes the need for this additional 
integration.  A final approach is to modify the definition
of $\hBs$ to allow equivalent DAG models to have
different parameter priors.


\section*{Acknowledgments}

We thank Chris Meek for helping us shape the definition
of DAG models and correcting earlier versions of this
manuscript, Bo Thiesson for implementing the proposed scheme,
and Jim Kajiya for his help in regard to the characterization
theorems.  
We also thank J\'{a}nos \Aczel,
Enrique Castillo, Clark Glymour,
Antal \Jarai,
Peter Spirtes, and the reviewers, for their useful suggestions.

\bibliographystyle{apalike}


\end{document}